\documentclass[10pt,twocolumn,letterpaper]{article}

\usepackage[pagenumbers]{cvpr} %

\usepackage{graphicx}
\usepackage{amsmath}
\usepackage{amssymb}
\usepackage{booktabs}
\usepackage{mathtools}

\usepackage[pagebackref,breaklinks,colorlinks]{hyperref}

\usepackage[capitalize]{cleveref}
\crefname{section}{Sec.}{Secs.}
\Crefname{section}{Section}{Sections}
\Crefname{table}{Table}{Tables}
\crefname{table}{Tab.}{Tabs.}

\def\cvprPaperID{4404} %
\def\confName{CVPR}
\def\confYear{2022}

\usepackage{overpic}
\usepackage{enumitem} %
\usepackage{overpic} %
\usepackage{color}
\usepackage{microtype} %

\definecolor{turquoise}{cmyk}{0.65,0,0.1,0.3}
\definecolor{purple}{rgb}{0.65,0,0.65}
\definecolor{dark_green}{rgb}{0, 0.5, 0}
\definecolor{orange}{rgb}{0.8, 0.6, 0.2}
\definecolor{red}{rgb}{0.8, 0.2, 0.2}
\definecolor{darkred}{rgb}{0.6, 0.1, 0.05}
\definecolor{blueish}{rgb}{0.0, 0.3, .6}
\definecolor{light_gray}{rgb}{0.7, 0.7, .7}
\definecolor{pink}{rgb}{1, 0, 1}
\definecolor{greyblue}{rgb}{0.25, 0.25, 1}

\newcommand{\tf}{\mathbf}

\newcommand{\loss}[1]{\mathcal{L}_\text{#1}}

\usepackage{blindtext}

\renewcommand{\paragraph}[1]{\vspace{1em}\noindent\textbf{#1}.}

\usepackage{microtype}

\begin{document}

\title{Deformable Sprites for Unsupervised Video Decomposition}

\author{
$\text{Vickie Ye}^1 \qquad
\text{Zhengqi Li}^2 \qquad
\text{Richard Tucker}^2 \qquad
\text{Angjoo Kanazawa}^1 \qquad
\text{Noah Snavely}^2$
\\
$^1\text{University of California, Berkeley} \qquad ^2\text{Google}$
}

\twocolumn[
\begin{@twocolumnfalse}
{%
\maketitle
\vspace{-3mm}
\centering
\includegraphics[width=\textwidth]{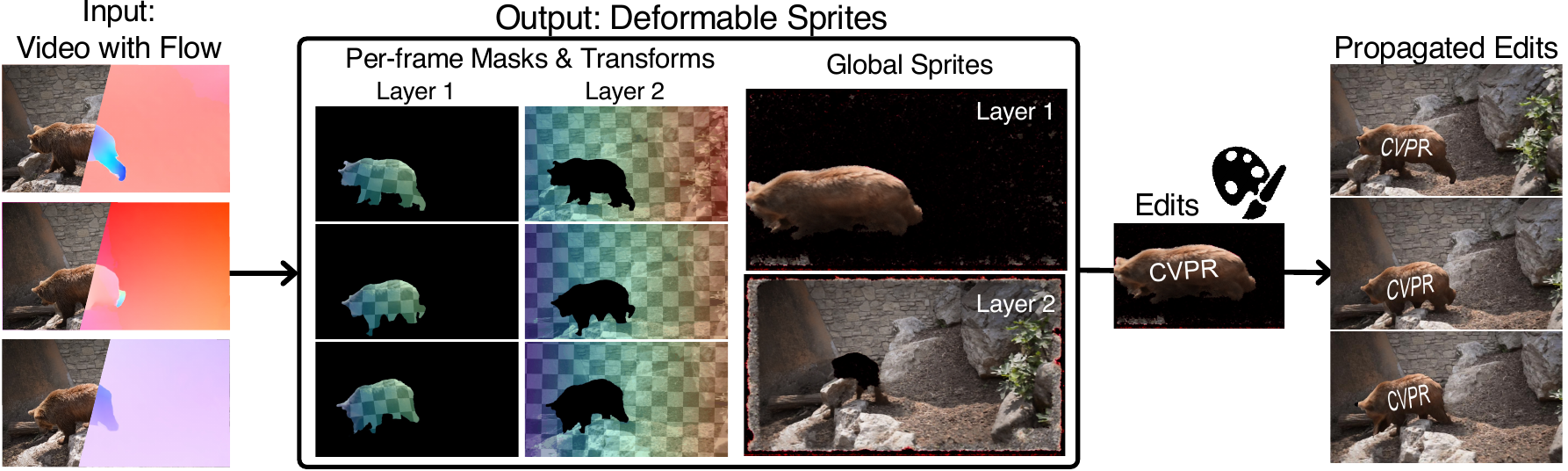}
\captionof{figure}{
	\textbf{Video Decomposition with Deformable Sprites.} 
	Given an RGB video and its optical flow (left), we present a method that decomposes the video into layers of persistent motion groups. 
	We represent each group as a \textit{Deformable Sprite} (center), which consists of an RGB sprite image, and masks and non-rigid transforms mapping the sprite to each frame. 
	We fit Deformable Sprites to each input video independently without any pre-training or user input. 
	The resulting decomposition captures long-term correspondences of sprites over time, enabling effects such as propagating edits on the sprite across the entire video (right).
	We show full videos of our results at the \href{https://deformable-sprites.github.io/}{project site}.
}
\label{fig:teaser}
\vspace{3mm}
}
\end{@twocolumnfalse}]

\begin{abstract}
We describe a method to extract persistent elements of a dynamic scene from an input video. 
We represent each scene element as a \emph{Deformable Sprite} consisting of three components: 
1) a 2D texture image for the entire video,
2) per-frame masks for the element,
and
3) non-rigid deformations that map the texture image into each video frame.
The resulting decomposition allows for applications such as consistent video editing. 
Deformable Sprites are a type of video auto-encoder model that is optimized on individual videos, and does not require training on a large dataset, nor does it rely on pre-trained models.
Moreover, our method does not require object masks or other user input, 
and discovers moving objects of a wider variety than previous work.
We evaluate our approach on standard video datasets and show qualitative results on a diverse array of Internet videos. 
\end{abstract}
\vspace{-1mm}
\section{Introduction}
\label{sec:intro}
When we observe a video of a dynamic scene, such as the bear video in Fig.~\ref{fig:teaser}, we do not see a disjoint set of pixels over time, but rather a bear walking in a zoo.
However, computer vision methods often represent videos as 3D raster pixel grids.
While this low-level representation is convenient for processing on hardware, it does not capture 
our intuitive notion of high-level objects moving through a 3D scene.

\emph{Moving layers} are an alternative representation proposed in the seminal work of Wang \& Adelson, in which scene elements are modeled as persistent image layers that transform over time to compose each video frame~\cite{wang1994representing}. 
Such layered representations capture the idea that there are persistent motion groups that move smoothly in the scene, while still accounting for sharp edges that result from occlusions. However, these classic methods were limited by machinery of the time to relatively simple motions and scenes.

Inspired by these classic ideas~\cite{wang1994representing,jojic2001learning,ravacha2008unwrap}, we present a new approach that decomposes videos of complex dynamic scenes into sets of persistent motion groups.
We do so by introducing \emph{Deformable Sprites} (Fig.~\ref{fig:teaser}, center),
representation of
motion groups across an entire video.
A Deformable Sprite for a motion group consists of three key components:
1) a canonical texture image, or \textit{sprite}, describing the group's appearance over all input frames, 
2) masks locating the group in each input frame,
and
3) a %
non-rigid geometric transformation that maps each sprite into each frame.
The resulting decomposition captures the correspondences of each motion group across the entire video,
such that modifications of the sprite can be propagated %
consistently throughout the video (Fig.~\ref{fig:teaser}, right).

We achieve this decomposition by fitting the Deformable Sprites representation to a video without any user input or even an \textit{a priori} notion of what kind of objects will be present.
Instead, the decomposition is derived solely from image and motion cues present in the video. Our approach optimizes the Deformable Sprites for each video independently, and does not require training on a dataset. This freedom from the need for training data allows our method to handle videos of novel objects and categories that are not labeled in standard segmentation benchmarks.

Our approach can handle videos with moving camera and articulated or deformable objects. To capture such non-rigid motion, we model the transformation as composition of a homography with 2D spatial splines that evolve smoothly over time. This explicit parameterization has the benefit of modeling non-rigid deformation with relatively few parameters while being continuous. The sprites and the masks are optimized through a convolutional neural network.

Our method has several advantages over recent approaches for video decomposition, which do not recover persistent layer appearances~\cite{yang2021dystab,lu2021omnimatte}, or require user inputs in the form of segmentation masks~\cite{lu2020Layered}.
Although our method outputs a rich video decomposition, and not simply object masks, we evaluate it on standard video object segmentation benchmarks where we obtain competitive results. On DAVIS~\cite{perazzi2016davis}, we obtain decomposition results that are similar to recent approaches that require user mask initialization~\cite{kasten2021layered}, while being faster to optimize (30 minutes vs.\ 10 hours) due to the low dimensionality of our deformation model. 
We 
further demonstrate our approach on a variety of Internet clips, where off-the-shelf segmentation methods do not generalize to discover meaningful groupings. To our knowledge, we present the first work that demonstrates video decomposition with a global texture model on in-the-wild videos without any user supervision. %

\begin{figure*}
\centering
\includegraphics[width=\linewidth]{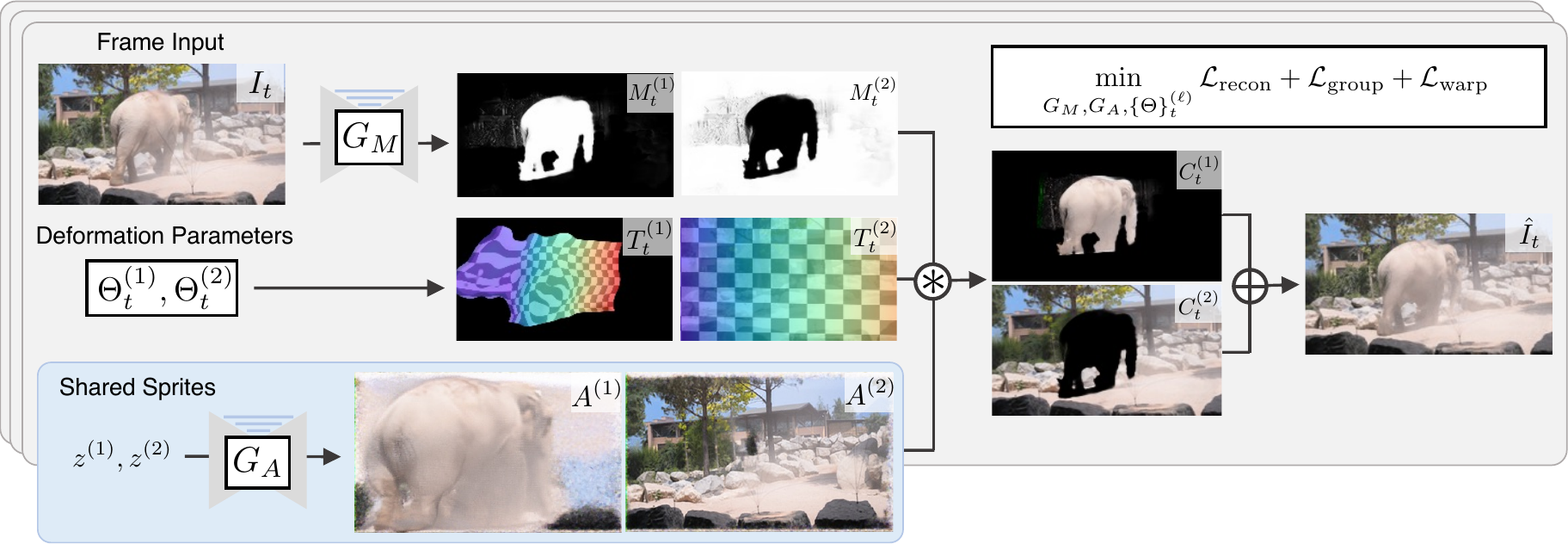}
\caption{
\textbf{Deformable Sprites,} a representation for persistent motion groups in a video of a dynamic scene. 
Here we represent $L=2$ motion groups: the elephant and background.
Deformable Sprites disentangle video into groups of three components: 
1) a single global appearance model, in the form of canonical per-layer RGB texture (\textit{sprites}) used across the entire video (denoted as $A^{(\ell)}$), 
2) per-frame masks indicating each group (denoted as $M_t^{(\ell)}$), 
and 3) spatio-temporal non-rigid transformations that capture the deformation and camera motion in each frame (denoted as $T_t^{(\ell)}$).
Transforming 
then masking the sprites results in a set of layers that are composited to reconstruct the input frame $t$ while preserving occlusion effects.
Deformable Sprites are optimized by minimizing per-frame losses on the reconstructions, masks and transformations described in Sec~\ref{sec:losses}.
}
\label{fig:pipeline}
\end{figure*}

\section{Related work}
\label{sec:related}

\noindent \textbf{Layer-based video decomposition.} Our work is inspired by the rich history of 
work that factorizes appearance and motion in videos by representing video frames as compositions of moving layers or sprites. 
Wang \& Adelson introduced this concept to computer vision in the 1990's~\cite{wang1994representing}, building on earlier work on estimation of multiple motions in image sequences~\cite{darrell1991robust,black1991robust}. Wang \& Adelson solve for a set of ordered RGB$\alpha$ layers with associated affine transformations using optical flow as a motion cue, and a number of other layer-based  decomposition methods followed~\cite{brostow1999motion,jojic2001learning,wills2003what,kumar2008learning}. These methods generally factor image sequences into appearance and motion according to different motion models (e.g. rigid, affine). 
Closely related %
is unwrap mosaics~\cite{ravacha2008unwrap}, which model objects in a video with a single global texture plus per-image warp fields and occlusion masks.

More recently, layered video decomposition methods extend these formulations to handle more complex, realistic videos. 
For instance, Lao et al. \cite{lao2018extending} extends the layered model to explicitly handle 3D motion.
More modern methods utilize neural networks in various ways, such as 
augmenting object masks~\cite{lu2021omnimatte}
or computing texture atlases and uv-coordinates for multiple layers using coordinate-based neural networks~\cite{kasten2021layered}.
One shared aspect of these works is that they require reasonable input masks of an object of interest, usually provided by the user or by segmentation networks trained on pre-defined object categories.
In contrast, we explore this problem in a fully unsupervised setting.

Our work is also closely related to MarioNette~\cite{smirnov2021marionette}, which decomposes a video into set of static sprites, and like us operates in an unsupervised manner. However, their learned sprites are static, not deformable, and so 
they mainly demonstrate their work on video game imagery,
where even animated sprites are just repetitions of a few discrete frames.
They must learn multiple representations of a single object to account for such motion. In contrast, we can learn a single global sprite per object, and model complex object motion using non-rigid transformations. Other recent work explores 3D object discovery in various settings~\cite{ost2021neural,yu2022unsupervised}, but can require manual annotation or simplified appearance models.

\medskip
\noindent \textbf{Motion segmentation.} As one aspect of computing our video representation, we also solve a motion grouping problem.
Approaches to motion segmentation include treating the problem as one of a spatio-temporal image clustering~\cite{shi1998motion}, or alternatively starting from motion estimates in the form of optical flow or point trajectories, and solving a grouping problem to associate pixels into a number of motion clusters~\cite{brox2010object,ochs2011object,ochs2014segmentation,keuper2015motion,keuper2017higher}. 
Recent approaches have also explored optimizing neural networks to map image and/or flow inputs to segmentation masks~\cite{xie2019object,dave2019towards}. 
Yang \etal, recognizing that the motion of a foreground object should not be predictable from background motion (and vice versa), propose an adversarial approach to compute segmentations that are mutually uninformative~\cite{yang2019unsupervised}.
While these works combine motion and appearance cues for segmentation,
they are designed for frame-by-frame prediction.
In contrast, our work computes explicit global texture images for each scene element, shared over all frames of a video.
Our resulting representation thus gives us long-term correspondences across entire videos.

\medskip
\noindent \textbf{Unsupervised video object segmentation.} 
While motion segmentation may or may not result in groups corresponding to semantic objects;
\emph{video object segmentation} (VOS) methods seek to detect objects in videos. 
Such methods are often trained on videos in advance---either in a fully supervised manner on datasets like DAVIS~\cite{perazzi2016davis}, or via self-supervision---to learn what constitutes objects in videos. 
Relevant to our work are approaches that do not consider full supervision. Such methods include that of 
Koh \etal~\cite{koh2017primary} and Hu \etal~\cite{hu2018unsupervised}, which leverage low- and mid-level cues like flow and edges to compute foreground objects (where the latter requires training an edge detector), and 
Yang \etal~\cite{yang2021selfsupervised}, which uses slot attention~\cite{locatello2020objectcentric} and self-supervision to learn to detect objects from optical flow.
DyStaB learns object saliency models using object motion in video, which can then be applied at test time to segment objects, even in static images~\cite{yang2021dystab}. 
In contrast to these approaches, we do not seek to detect semantic object categories, but to find the decomposition that best explains the scene motion and its resulting appearance.
As we show below, this allows us to represent moving elements that do not fall into traditional object categories.

\section{Deformable Sprites}

\newcommand{\Ft}{F_{t\rightarrow t+1}}

The inputs to our method are video frames of a dynamic scene $\{I_t \in \mathbb{R}^{H\times W}\}_{t=1}^N$, and optical flow computed between consecutive frames, $\{ \Ft \}_{t=1}^{N-1}$.
From these inputs, we aim to recover a set of $L$ distinct \emph{motion groups} in the dynamic scene.
We want these groups to capture the underlying elements in the video, moving smoothly through the scene.
To do this, we introduce \emph{Deformable Sprites} to consistently represent motion groups as they evolve throughout a video.
A Deformable Sprite consists of 
a canonical appearance model in the form of 
a texture image, or \emph{sprite}, for each group, used to describe the motion group's appearance across all video frames,
and geometric transformation and mask models, used to relate the motion group's position and geometry in each input frame to the sprite.

We illustrate our representation, and our approach to fitting it to an input video, in Fig.~\ref{fig:pipeline}.
For each frame $I_t$, we predict a mask for each of the $L$ motion groups, denoted $\{ M_t^{(\ell)} \}_{\ell=1}^L$.
In the figure, there are two masks corresponding to two motion groups, elephant and background. 
Each motion group 
has its own RGB texture sprite describing the group's appearance across the entire video.
We denote these $L$ texture images as $\{ A^{(\ell)} \}_{\ell=1}^L$.
Finally, we need to know where to place and how to deform each sprite in each frame, \eg to match the position and pose of the elephant in the figure.
To that end, we estimate a spatio-temporal spline-based transformation $T_t^{(\ell)}$ between the canonical texture coordinates and the frame coordinates of $I_t$, for every motion group $\ell$ and every frame $t$.
Given a set of Deformable Sprites, 
we can reconstruct any video frame $I_t$ by using the associated transformations to warp the global textures into the given frame, masking each layer according to the computed masks, then composing the layers into a single RGB image.
This yields a reconstructed image $\hat{I}_t$.

\medskip
\noindent \textbf{Design motivations.} 
A natural way to fit any kind of layered model to images is to simply optimize the layers and transforms to minimize a reconstruction loss.
However, this video decomposition problem is severely ill-posed---a large variety of layer decompositions might perfectly reproduce the video but yield nonsensical group separations.
We seek a natural decomposition where the discovered layers correspond to objects that are coherent within a frame and consistent across time.
This desire for a coherent decomposition suggests that the decomposition should be compact and low-dimensional, which is a key motivation for representing the appearance of each motion layer with a fixed sprite texture. 
As a consequence, %
we must also represent the geometry of each sprite per frame---where to place it, %
and how to pose it. We employ a spatio-temporal spline to achieve this non-rigid transformation, which can be parameterized with a small number of explicit parameters.

\subsection{Grouping}
A key question in computing our layered representation from a video is: for each frame, which pixels belong to which layer?
Many prior works use optical flow cues to compute explicit motion groups, either with robust parametric model fitting (e.g. \cite{wills2003what,wang1994representing}), or with robust clustering methods (e.g. \cite{ochs2011object,ochs2012higher,ochs2014segmentation,keuper2017higher,liu2005motion}).
However, explicit techniques struggle to handle objects with multiple motions and complex trajectories.
Moreover, we are not solely interested in computing motion groups per-frame, but rather in representing the both the appearance and the motion of a group as it evolves through the entire video.
For these reasons,
we opt to learn a single CNN grouping model, $G_m$, for each video sequence (and shared across all frames of that sequence),
to predict masks that are consistent with appearance cues.
$G_m$ takes in an RGB image $I_t$ and predicts $L$ soft masks
$\{B_t^{(\ell)}\}_{\ell=1}^L \in [0, 1]^{H\times W}$ for that frame, one per group.
We then use back-to-front compositing to convert these to probability masks that sum to 1 (per pixel) across the groups:
\begin{equation}
    M_t^{(\ell)} = B_t^{(\ell)} \cdot \prod_{i=\ell + 1}^{L} (1 - B_t^{(i)}).
\end{equation}
Our final grouping is the per-pixel mask, $M_t^{(\ell)}$, for each layer $\ell$ and video frame $t$.
We use optical flow in computing motion groups;
however, rather than explicitly computing groups directly from flow,
we use it in our losses as supervision (Sec.~\ref{sec:losses}),
and let the model learn which appearance cues in the input frames to associate with each motion group.

\subsection{Global Sprite Images}
The central component of Deformable Sprites is a global sprite image that represents each layer's appearance across the entire video as a fixed RGB texture.
In the case of a dynamic foreground object, the canonical representation helps group together parts of the object moving in complex ways.
For example, consider the walking bear in Fig.~\ref{fig:teaser}. %
As it walks, different parts of the body undergo distinct motions.
These parts are difficult to group together if one looks at motion alone.
However, a global sprite image can group these distinct trajectories together because the collective body has a consistent appearance over time.

For each layer $\ell$, we represent its appearance across the entire video as a fixed texture image $A^{(\ell)}\in \mathbb{R}^{H'\times W'}$. 
We leverage the natural image prior of CNNs~\cite{ulyanov2020deep} and optimize each sprite image $A^{(\ell)}$ through a shared CNN, $G_A$, with a fixed code $z\in \mathbb{R}^{H'\times W'\times d}$ uniformly sampled from $[0, 1]$.

While the global sprites provide cues to group parts that move differently in a single frame, a single, fixed sprite is not sufficient if the camera or the object moves in the scene. Therefore we estimate a set of transformations that can non-rigidly deform the sprites to match the appearance of the motion groups in each frame. 

\subsection{Spatio-Temporal Deformation with Splines}
In general, objects move about the world in a smooth fashion. 
However, as noted by 
Wang \& Adelson~\cite{wang1994representing}, modeling the 2D motion field as a smooth function fails to capture sharp edges resulting from occlusion boundaries. 
The benefit of a layered representation like ours, 
then, is that the masks capture the visibility and occlusion boundaries, allowing the transformations to be modeled as smooth functions.

Specifically, we model the motion of each sprite with a transformation that maps
the pixel coordinates of frame $t$ to the texture coordinates of layer $\ell$.
We represent this transformation as a \textit{continuous} function $T^{(\ell)}(t)$ that is smooth over space and time via two splines, one for space and one for time. %
This approach has the benefit of modeling non-rigid deformation with relatively few motion parameters. For simplicity, we omit layer superscripts in the notation below. %

\medskip
\noindent \textbf{Splines in space.}
The spatial transformation evaluated at time $t$ is
modeled as a combination of rigid and non-rigid 2D transformations $\tf\Theta(t)=[\eta(t),\tf V(t)]$ . 
The rigid component is modeled as a homography $\tf H(t)$, parameterized by 8 parameters $\eta(t)\in \mathbb{R}^8$. 
This component captures scaling effects and simple object and camera motions.
We assume that the background layer, assigned to layer index $L$, is largely static (and hence only moving due to camera motion), and so we only model its motion with a homography, without a non-rigid component.

The non-rigid component of the motion is modeled by a 2D B-spline, which defines a smoothly varying deformation field $D$.
The B-spline is parameterized by a grid of fixed control points
$X \in [-1, 1]^{N_i \times N_j \times 2}$,
with uniform spacing $\Delta_{\tf x}$ in the input (view) frame coordinates.
We then define a grid of parameters, $\tf{V}(t) \in \mathbb{R}^{N_i \times N_j \times 2}$, at these control points.
These parameters describe the position of the frame control point when it is transformed into texture (canonical) coordinates.
Then at a point $\bf x$ in the input frame, the deformation vector $\tf{v} = D(\tf{x}; \tf{V})$ can be interpolated as
\begin{equation}
    D(\tf{x}; \tf{V}) = \sum_{i, j} \tf{V}_{i,j} \cdot B^{\text{2d}}_d\Bigg(\frac{\tf{x} - X_{i,j}}{\Delta_{\tf{x}}} \Bigg),
\end{equation}
where $B^{\text{2d}}_d$ is a degree $d$ piecewise polynomial defined over the unit square $[0, 1]^2$; we use degree 2.
More details about B-spline interpolation are provided in the supplemental.

\medskip
\noindent \textbf{Keyframes over time.}
We expect corresponding points within each layer to deform smoothly over time.
To enforce this smoothness, we represent our transformations over time as a 1D function parameterized by temporal control points or \textit{keyframes}.
This allows us to represent temporally varying transformations with a smaller number of parameters compared to estimating a dense transformation separately for every time $t$ while ensuring smoothness.
We store a set of $N_T$ keyframes of transformation parameters $\{\tf \Theta_i\}_{i=1}^{N_T}$ at times $\{t_i\}_{i=1}^{N_T}$, for every layer. %
The keyframes are spaced uniformly at time intervals $\Delta_T := \frac{N}{N_T}$.
At time $t$, we interpolate between these keyframes with a 1D B-spline~\cite{bartels1995introduction}:
\begin{equation}
    \tf \Theta(t) = \sum_{i} \tf \Theta_i \cdot B^{\text{1d}}_d\Bigg(\frac{t - t_i}{\Delta_T}\Bigg),
\end{equation}
where $B^{\text{1d}}$ is a degree $d$ piecewise polynomial defined over the unit interval $[0, 1]$; we use degree 2.

The final transformation from an image coordinate $\bf x$ to texture coordinate $\bf p$ of layer $l$ at time $t$ is then
\begin{equation}
  \tf{p} = T^{(\ell)}_t(\tf{x}) = H(\tf x; \tf \eta^{(\ell)}(t)) + D(\tf x; \tf V^{(\ell)}(t)),
\end{equation}
where $\tf{\eta}^{(\ell)}(t)$ and $\tf V^{(\ell)}(t)$ are interpolated values at time $t$. %
The parameters to be estimated are %
the spatio-temporal control points $\{\tf \Theta_i\}_{i=1}^{N_T}$, where $ \tf\Theta_i := [\tf{\eta}_i, \tf{V}_i]$.

\medskip
\noindent \textbf{Compositing.}
Given a set of Deformable Sprites representing a video, we can reconstruct a frame $\hat{I}_t$ at time $t$ as follows.
For each layer $\ell$, we resample the canonical texture $A^{(\ell)}$ with the transformation $T_t^{(\ell)}$
into texture coordinates to get frame appearance $C_t^{(\ell)}$.
Then, each appearance can be composited with the masks $\{M_t^{(\ell)}\}_{t=1}^N$
to get our final reconstruction.

\section{Training Deformable Sprites}
\label{sec:losses}
We fit our Deformable Sprites model using gradient descent to optimize model parameters, namely the per-frame, per-layer masks $M_t^\ell$, per-layer spatio-temporal transformation
keyframes $\{\tf \Theta_i\}_{i=1}^{N_T}$, and per-layer global appearance sprites $A^\ell$.
Our primary loss for optimizing the representation is a video reconstruction loss. %
However, because our recovery problem is under-determined,
we add regularizing losses to encourage our model toward solutions with sprites that (1) move coherently,
and (2) are consistent over time.

\begin{figure*}[th]
\centering
\includegraphics[width=\linewidth]{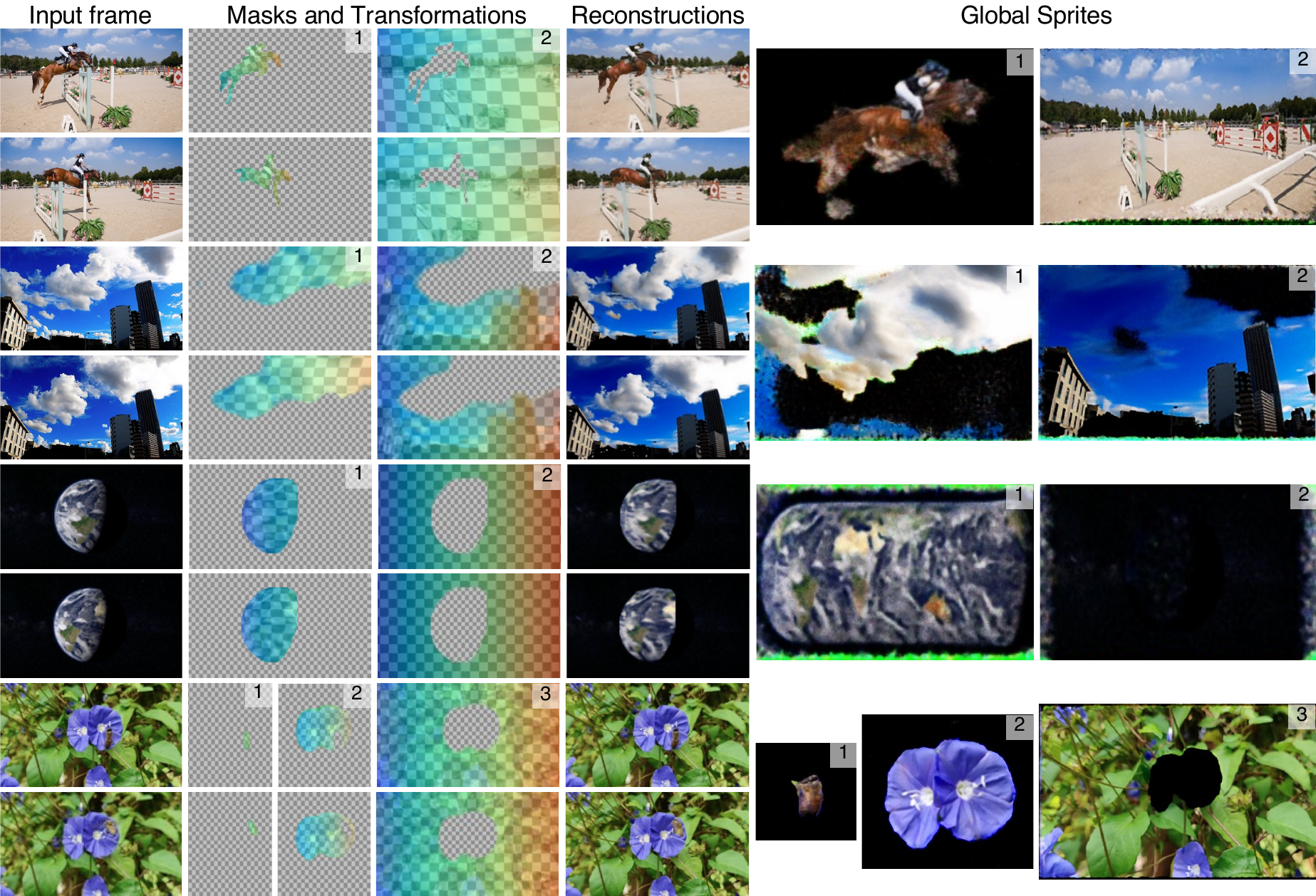}
\caption{
\textbf{Qualitative results on diverse videos.}
We show examples with two and three (second row) motion groups and examples with non-traditional foreground objects (bottom two rows).
We show two input frames for each example.
For each frame, we show the masks and transformations for each motion group (where the number in the corner of the image indicates the group index), as well as the composited reconstruction.
Finally, we show the global sprites for each motion group, shared between all frames (with index similarly indicated).
We overlay a rainbow checkerboard on the masks to visualize the corresponding sprite texture coordinates.
Note for instance the decomposition into bee, flower, and background in the second video, the layer that captures non-object-like cloud regions in the third video, and the aggregation of the globe texture across the full video in the last video. We show full video results and more examples in the supplemental.
}
\label{fig:qualitative_all}
\end{figure*}

\medskip
\noindent \textbf{Motion grouping loss.}
We determine how to group together pixels using losses on the optical flow vectors between consecutive frames.
In particular, we use the optical flow field to estimate the static background content, as well as to group together dynamic pixels that move similarly to each other:
\begin{equation}
    \loss{group} = \loss{static} + \loss{dynamic}.
\end{equation}
We denote the optical flow between frames $t$ and $t+1$ as $\Ft$;
we refer to flow correspondences as $(x, x')$, where $x' = \Ft(x)$.

$\loss{static}$ uses a robust estimate of the two-view geometry to separate static from dynamic points.
We estimate the fundamental matrix between frames $t$ and $t+1$, $\hat{\tf F}_{t\rightarrow t+1} \in \mathbb{R}^{3\times 3}$, with least median of squares (LMedS) regression~\cite{rousseeuw1984least} on the dense correspondences from flow.
Note that we fit $\hat{\tf F}_{t\rightarrow t+1}$ to all pixels in the frame, and must detect when the LMeDS estimate fails capture the static geometry, when non-static pixels outnumber static pixels (see the supplement).

We then compute the Sampson distance $\varepsilon_t(x, x')$ using the estimated fundamental matrix \cite{hartley2003multiple}.
The Sampson distance approximately captures how well the correspondence $(x, x')$ adheres to the epipolar geometry of $\hat{\tf F}_{t \rightarrow t+1}$;
points with high Sampson distance are likely moving, and therefore do not obey the geometric constraints.
We thus penalize the background layer with
\begin{equation}
     \loss{static} = \textstyle\sum_x
     \varepsilon_t(x) \cdot M_t^{(L)}(x)
    + \beta \cdot \big(1 - \varepsilon_t(x)\big) 
    \cdot \big( 1 - M_t^{(L)}(x) \big),
\end{equation}
where $M_t^{(L)}$ is the background mask, and $\beta=0.002$.

We further wish to group dynamic points that move similarly to each other.
We encourage this with $\loss{dynamic}$, which minimizes the distortion of the optical flow in each moving group, much like in k-means clustering.
For each group $\ell$, we determine a group exemplar flow vector $\mu_t{(\ell) \in \mathbb{R}^2}$.
We then penalize the group assignments with
\begin{equation}
    \loss{dynamic} = \textstyle\sum_{\ell, x} M_t^{(\ell)}(x) \cdot \|\Ft(x) - \mu_t^{(\ell)} \|^2.
\end{equation}
We let the exemplar for each layer $\ell$ be the weighted mean of the flow vectors assigned to that layer:
\begin{equation}
    \mu_{t}^\ell \leftarrow \big(\textstyle\sum_{x} M_{t}^\ell(x) \cdot \Ft(x) \big) \big/ \big(\textstyle\sum_{x} M_{t}^\ell(x)\big).
    \label{eq:mean_flow}
\end{equation}

\medskip
\noindent \textbf{Optical flow consistency losses.}
Ideally, even as points in the scene move with respect to the camera, they should map to the same coordinates in the texture map.
To encourage this behavior, we penalize inconsistencies between the mappings, masks, and optical flow for each layer:
\begin{equation}
    \loss{warp} = \loss{transform} + \loss{mask}.
\end{equation}
We encourage the transformations to be consistent with the flow between consecutive frames $t$ and $t+1$. 
We also make the loss scale-invariant to stabilize training and prevent degenerate transforms.
Hence we formulate our loss as
\begin{equation}
    \loss{transform} = \sum_{\ell, x} M_{t}^\ell(x) \cdot  \frac{\| T_t^\ell(x) - T_{t+1}^\ell(x') \|}{s_t^{(\ell)} + s_{t+1}^{(\ell)}} 
\end{equation}
where $x' = \Ft(x)$, and $s_t^{(\ell)}$ is the scale of $T_t^{(\ell)}$ (see the supplement for details).

Likewise $\loss{mask}$ encourages masks to be consistent with flow: 
\begin{equation}
    \loss{mask} = \textstyle\sum_{\ell, x} \| M_{t}^\ell(x) - M_{t+1}^\ell(x') \|.
\end{equation}

\smallskip
\noindent \textbf{Reconstruction loss.}
We use a combination of L1 distance and Laplacian pyramid similarity as our reconstruction loss:
\begin{equation}
    \loss{recon} = \| \hat{I}_t - I_t \|_1 + \loss{edges},
\end{equation}
with
$
    \loss{edges} = \sum_{m} 4^m \| \textrm{Lap}_m(\hat{I}_t) - \textrm{Lap}_m(I_t) \|_1, $%
where $\textrm{Lap}_m$ is the $m$-th level of the Laplacian pyramid.%

The final loss function is then:
\begin{equation}
    \mathcal{L} = \lambda_{\text{recon}} \loss{recon}
    + \lambda_{\text{group}} \loss{group}
    + \lambda_{\text{warp}} \loss{warp}.
\end{equation}

\noindent \textbf{Implementation details.}
\label{sec:impl}
We compute optical flow with RAFT \cite{teed2020raft}.
We use UNets \cite{ronneberger2015unet} for the mask prediction and texture generation models.
We use $\Delta_T = 4$ and $N_i = 16, N_j = \lfloor \frac{16 * W}{H} \rfloor$ for the spline knots.
We introduce losses to the optimization problem in a schedule: we first warmstart the grouping network with $\loss{group}$.
We use these initial rough masks to initialize the scales and translations
of the frame-to-texture transforms for each layer, then add $\loss{warp}$ and $\loss{recon}$.
Please refer to the supplemental for additional details.
Code can be found at the \href{https://deformable-sprites.github.io/}{project website}.

\begin{figure}[t]
\centering
\includegraphics[width=\linewidth]{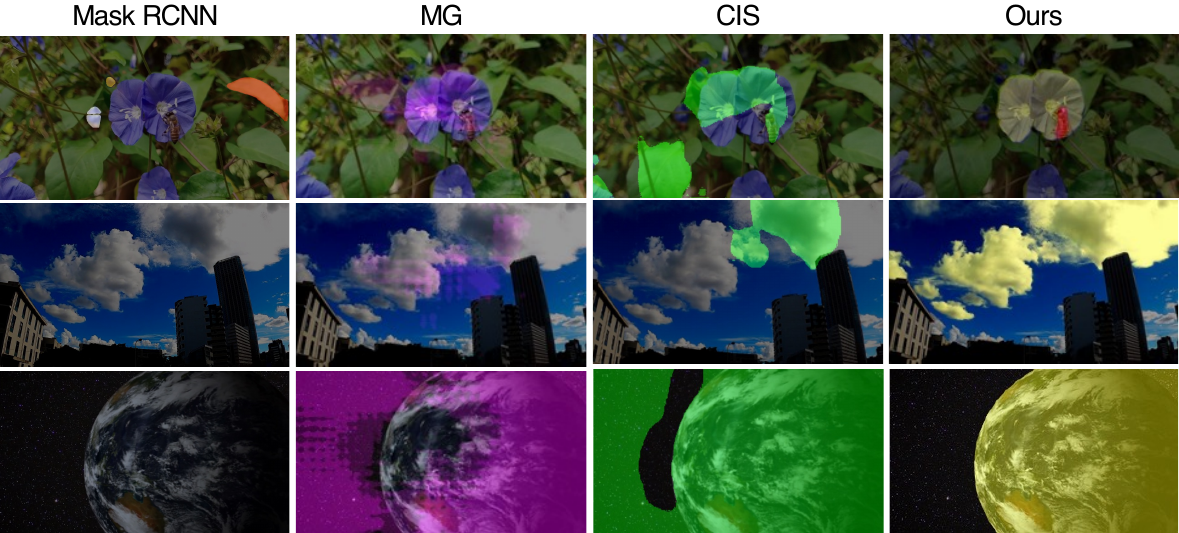}
\caption{
\textbf{Segmentation models on non-traditional objects}. 
We show predicted segmentations on frames of the bottom three video sequences in Figure~\ref{fig:qualitative_all}.
We show the masks from MaskRCNN \cite{he2018mask}, trained on COCO categories, and two recent motion segmentation methods, MG \cite{yang2021selfsupervised} and CIS \cite{yang2019unsupervised}, trained on DAVIS\cite{perazzi2016davis}.
These methods are unable to handle the out-of-distribution images.
}
\vspace{-2mm}
\label{fig:maskrcnn_fails}
\end{figure}

\begin{figure}[t]
\centering
\includegraphics[width=\linewidth]{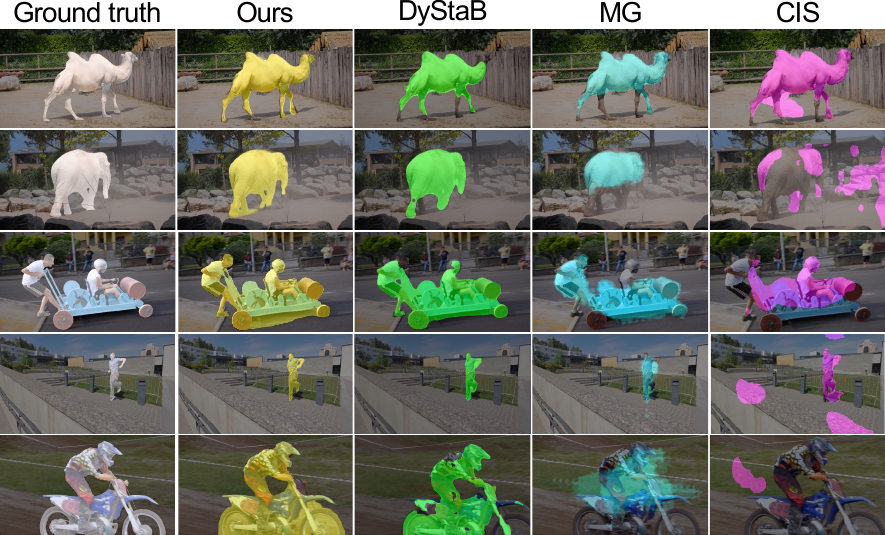}
\caption{
\textbf{Qualitative mask comparison with top baselines on DAVIS}.
We compare the masks from Deformable Sprites with baselines on DAVIS\cite{perazzi2016davis}.
We compare with DyStaB~\cite{yang2021dystab}, MG~\cite{yang2021selfsupervised}, and CIS~\cite{yang2019unsupervised}.
Please see video results in the supplemental.
}
\vspace{-2mm}
\label{fig:davis}
\end{figure}

\section{Results}

\begin{table}
\centering
\begin{tabular}{@{}lccc@{}}
\toprule
& DAVIS & FBMS & SegTV2 \\
\midrule
ARP \cite{koh2017primary}
& 76.2 & 59.8 & 57.2 \\
ELM \cite{lao2018extending}
& 61.8 & 61.6 & - \\
MG \cite{yang2021selfsupervised}
& 68.3 & 53.1 & 58.2 \\
CIS \cite{yang2019unsupervised} 
& 71.5 & 63.5 & 62.0 \\
$\text{DyStaB}^*$ \cite{yang2021dystab} 
& 80.0 & 73.2 & 74.2 \\
\textbf{Ours}
& 79.1 & 71.8 & 72.1 \\
\bottomrule
\end{tabular}
\caption{
\textbf{Quantitative mask evaluation on VOS benchmarks.}
We compare masks from Deformable Sprites with top-performing baselines on the DAVIS~\cite{perazzi2016davis} and SegTrackV2~\cite{li2013video} benchmarks. We achieve IOU $(\mathcal{J})$ scores competitive with current SOTA.
$^*$(Note that DyStaB \cite{yang2021dystab} is trained on a video dataset; all other methods use only the information present in the input video.)
} %
\label{tab:davis}
\vspace{-3mm}
\end{table}

\subsection{Qualitative results on real videos}
In Figure~\ref{fig:qualitative_all}, we show our estimated Deformable Sprites for a variety of real videos.
The top row is from the DAVIS dataset \cite{perazzi2016davis}; the bottom three are Internet videos.
We show results for a video with three sprites (second row), in which both the flower and bee are moving foreground objects, as well as for videos with non-traditional foreground objects, such as clouds (third row) and Earth (bottom row).
We show result videos and additional examples in the supplemental.

For each example, we show two frames from the input video and our Deformable Sprites representation.
Each frame has masks and transforms corresponding to each sprite; we visualize the masks and transforms for each layer together.
We overlay a rainbow checkerboard on top of the masks to show corresponding sprite texture coordinates;
we see that the same texture coordinates follow the same points of the object throughout the video.
The sprite images are also completed as regions in each layer become visible over the course of the video.
The globe example in Figure~\ref{fig:qualitative_all} demonstrates this effect: as the globe rotates, the foreground sprite is completed into an unwrapped world map.

We recover our Deformable Sprites based solely on the motion and appearance cues present in the input, and do not rely on training data.
As such, we can easily recover Deformable Sprites for objects that do not fall into common categories.
This is in contrast to layer decomposition methods such as\cite{lu2021omnimatte}, \cite{ravacha2008unwrap} or \cite{kasten2021layered},
which rely on input masks, either from off-the-shelf pre-trained segmentation models or from user interaction.
In Figure~\ref{fig:maskrcnn_fails}, we show the masks we obtain from our representation compared to off-the-shelf Mask RCNN, and
two recent motion segmentation methods \cite{yang2019unsupervised}, \cite{yang2021selfsupervised}, trained on DAVIS.
Layer decomposition methods \cite{kasten2021layered}, \cite{lu2021omnimatte} that require input masks would require further user interaction, e.g., using methods such as GrabCut~\cite{rother2004grabcut} to decompose these scenes.

\begin{figure*}[h]
\centering
\includegraphics[width=\linewidth]{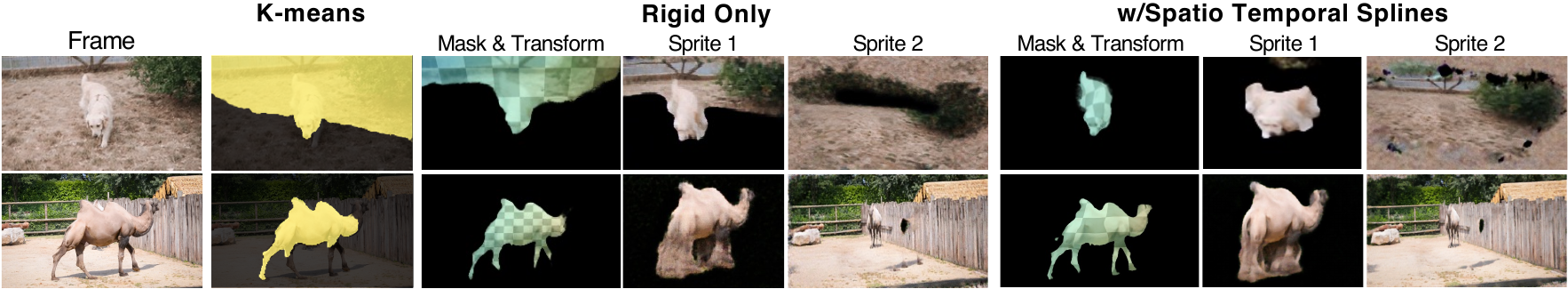}
\caption{
\textbf{Ablation.}
We ablate our model without global sprites or transformations and trained only with the motion grouping loss (\textbf{K-means}),
without only a rigid transformation model (\textbf{Rigid Only}), and compare with the full model (\textbf{w/Spatio Temporal Splines}).
Modeling non-rigid motion enables us to capture the twisting dog and the camel's legs.
}
\vspace{-3mm}
\label{fig:ablations}
\end{figure*}

\subsection{Motion segmentation results on benchmarks}
A key advantage of Deformable Sprites is the ability to discover reasonable motion groups during optimization.
In many cases, moving objects in a video are commonly segmented objects, such as animals, peoples, or cars.
We compare the quality of our motion groups to other unsupervised motion segmentation baselines on two standard video-object segmentation benchmarks, DAVIS2016 \cite{perazzi2016davis} and SegTrackV2 \cite{li2013video}.
We report quantitative performance in Table~\ref{tab:davis}, in which our method is the second best in both benchmarks.
We show qualitative comparisons with recent motion grouping methods DyStaB \cite{yang2021dystab}, CIS \cite{yang2019unsupervised} and MG \cite{yang2021selfsupervised} in Figure~\ref{fig:davis}.
Our representation can better capture the limbs of articulated animals and people, thanks to our spline representation.
Finally, we note that while motion group boundaries generally correspond with the semantic object masks, they also often include effects such as reflections and shadows.
Sometimes this behavior is desired~\cite{lu2021omnimatte}, although these effects are not included in the segmentation task we evaluate on.

\subsection{Ablations}
\begin{table}
\centering
\begin{tabular}{@{}lccc@{}}
\toprule
& k-means & rigid only & \textbf{+ splines (full)} \\
\midrule
DAVIS & 64.6 & 71.5 & 79.1\\
\bottomrule
\end{tabular}
\caption{
\textbf{Ablations}
We compare our full model with the variants \textit{k-means} and \textit{rigid only} on the DAVIS dataset. \textit{k-means} comprises of only a mask prediction network trained with $\loss{group}$ and $\loss{mask}$. \textit{rigid only} adds back the sprites, using only homographies as transforms.
}
\label{tab:ablations}
\vspace{-3mm}
\end{table}

We ablate the effect of the global appearance model and the deformable transforms on the quality of the output masks on the DAVIS dataset;
we report the quantitative differences in Table~\ref{tab:ablations}.
We remove sprite textures and transformations and train the mask prediction network with the motion grouping loss $\loss{group}$ and mask consistency loss $\loss{mask}$ (K-means);
the coherence of the masks for this variant drops dramatically.
We then add back the sprite textures, but only use homographies as transformations (rigid only).
We see in Figure~\ref{fig:ablations} that the persistent sprites help in recognizing and segmenting objects when they are stationary.
However, homographies cannot model long-term correspondences of deforming objects.
We see in Figure~\ref{fig:ablations} that the stationary camel's legs and dog's body are still grouped with the foreground, and the corresponding texture is more complete.

\subsection{Consistent video editing}
\begin{figure}[t]
  \centering
    \begin{tabular}{@{\hspace{-0.05em}}c@{\hspace{-0.05em}}c@{\hspace{-0.05em}}c@{\hspace{-0.05em}}}
    \includegraphics[width=0.32\columnwidth]{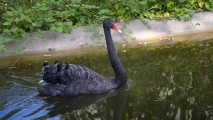} \vspace{-0.1em} &
    \includegraphics[width=0.32\columnwidth]{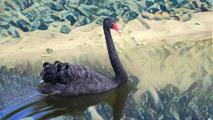} \vspace{-0.1em} &
    \includegraphics[width=0.32\columnwidth]{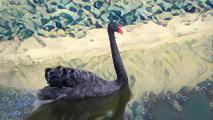} \vspace{-0.1em} \\ 
    \includegraphics[width=0.32\columnwidth]{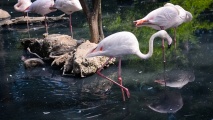} \vspace{-0.1em}&
    \includegraphics[width=0.32\columnwidth]{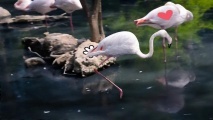} \vspace{-0.1em}&
    \includegraphics[width=0.32\columnwidth]{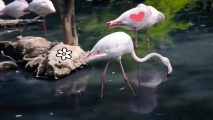} \vspace{-0.1em} \\
    \includegraphics[width=0.32\columnwidth]{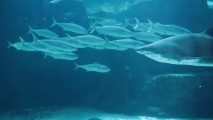} \vspace{-0.1em}&
    \includegraphics[width=0.32\columnwidth]{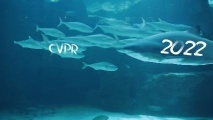} \vspace{-0.1em}&
    \includegraphics[width=0.32\columnwidth]{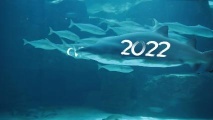} \vspace{-0.1em} \\ 
    {\small \text{Input frame}} \vspace{-0.1em} & {\small \text{Edited frame $1$}} \vspace{-0.1em} & {\small \text{Edited frame $2$}} \vspace{-0.1em}
    \end{tabular}
  	\caption{
  	\textbf{Consistent video editing.} We can directly edit the recovered textures; the edits are then automatically propagated to the full output video.
  	In the first row, the background is stylized; in the second two rows, decals are added to objects.
  	The last row has two moving foreground groups.}\label{fig:fg_bg_edit}
    \vspace{-3mm}%
\end{figure}

\begin{figure}[t]
  \centering
    \begin{tabular}{@{\hspace{-0.05em}}c@{\hspace{-0.05em}}c@{\hspace{-0.05em}}}
    \includegraphics[width=0.5\columnwidth]{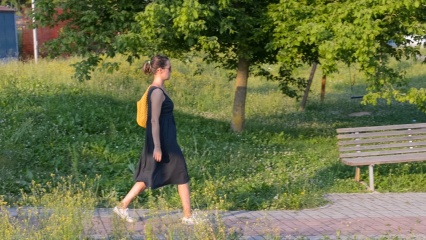} \vspace{-0.1em} &
    \includegraphics[width=0.5\columnwidth]{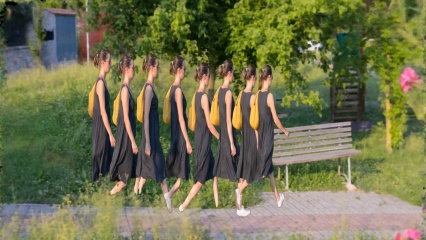} \vspace{-0.1em} \\
    \includegraphics[width=0.5\columnwidth]{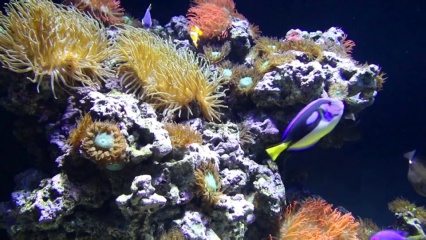} \vspace{-0.1em} &
    \includegraphics[width=0.5\columnwidth]{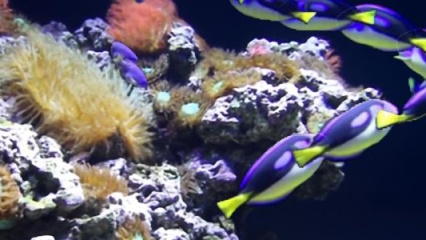} \vspace{-0.1em} \\
    {\small \text{Sample input frame}} \vspace{-0.1em} & {\small \text{Motion sculpture}} \vspace{-0.1em} 
    \end{tabular}
  	\caption{\textbf{Motion sculpture.} Our method can produce motion sculptures by compositing the foreground texture at several different times onto the background via the recovered transformations. }\label{fig:motion_sculpture}
    \vspace{-3mm}%
\end{figure}

Our method enables downstream applications such as  creating video effects. In particular, we demonstrate consistent video editing by directly editing our learned foreground and background texture atlases. 
In Figure~\ref{fig:fg_bg_edit}, we modify the sprite images with style transfer~\cite{zhang2017multistyle} or by adding decals, and consistently propagate edits with our learned texture mappings. 
Our method can also create 2D motion sculpture effects, as shown in Figure~\ref{fig:motion_sculpture}, by transforming foreground texture and learned alpha masks for several frames onto the background texture and overlaying them together. 
We refer readers to our \href{https://deformable-sprites.github.io/}{project website} for full video results.

\section{Discussion and Conclusion}

Our approach has a few limitations. One is that our fixed appearance model does not explicitly handle changes in appearance over time, for instance due to lighting (e.g., if a person walks through a region in shadow). 
Such appearance changes can be modeled in unintended ways, e.g., through clever uses of unused part of the texture or use of blending via the soft masks; adding explicit modeling of appearance changes would be an interesting extension of our method. 
Because the layer ordering is weakly constrained for non-background elements, when multiple foreground layers are present they may not end up in a natural order.
Finally, our method is limited to modeling elements with 2D textures and transformations; it would be interesting to extend our approach to true 3D decompositions of scenes.

In summary, we presented a new method for decomposing videos into layers that combines neural representations with classic ideas in video representation, and show the effectiveness of our method across a range of challenging videos, and for applications in video editing.

{
    \clearpage
    \small
    \bibliographystyle{ieee_fullname}
    \bibliography{macros,main}
}

\end{document}